\documentclass[11pt]{article}

\usepackage[margin=1in]{geometry}
\usepackage{amsmath,amssymb}
\usepackage{booktabs}
\usepackage{graphicx}
\usepackage{hyperref}
\usepackage{xcolor}
\usepackage{listings}
\usepackage{enumitem}
\usepackage{tikz}
\usetikzlibrary{arrows.meta,positioning,fit,calc,shapes.geometric}
\usepackage{caption}
\usepackage{subcaption}
\usepackage{microtype}
\usepackage{setspace}

\hypersetup{
  colorlinks=true,
  linkcolor=blue!60!black,
  citecolor=blue!60!black,
  urlcolor=blue!60!black
}

\title{On-Premises Multi-Course RAG Tutoring for Business Education:\\
Hardware--Software Trade-offs in a Campus AI Tutor}

\author{
  Sidney Shapiro\thanks{ORCID: \url{https://orcid.org/0000-0003-4193-1154}}
  \qquad Joshua Lindemann\\
  Dhillon School of Business, University of Lethbridge
}

\date{}

\begin{document}
\maketitle

\begin{abstract}
Campus AI tutors based on retrieval-augmented generation (RAG) must ground
answers in assigned course materials while keeping textbooks and student
dialogue on institutional infrastructure. We present CourseChat, an
on-premises, multi-course RAG tutor for undergraduate business education,
deployed behind a campus web gateway and intended for Moodle-embedded use.
Six isolated course offerings---each keyed by its own course reference number
(CRN)---share twin edge AI hosts running a FastAPI service, a local vector
database, and a local large language model (LLM) served by Ollama. We report
two generation-model bake-off rounds, a separate fixed-evidence source-fidelity
comparison, and conversation and quiz audits. Several larger models failed a
classroom speed gate, but a 12B model and a 7B alternative passed it. A
separate mixture-of-experts candidate improved some corrections while
introducing new factual and continuity errors. We therefore retain the 8B
production model pending a demonstrated overall improvement, rather than
claiming that 8B is universally optimal. Software changes improved follow-up
topic resolution while preserving course scope; 435 prebuilt questions across
65 modules decouple practice from live generation. The results support
treating model choice, evidence selection, serving compatibility, and product
design as a joint engineering decision. They do not establish learning gains:
faculty ratings, peak-load capacity, and complete public-gateway acceptance
remain separate evaluation needs.
\end{abstract}


\section{Introduction}
\label{sec:intro}

Campus AI tutors promise grounded help on assigned materials without sending
copyrighted textbooks or student dialogue to external APIs. Realizing that
promise under institutional constraints is a systems problem: limited GPU
memory, concurrent classroom use, LMS embeds, and a hard requirement that
Course~$A$ never answer from Course~$B$'s corpus.

This paper documents CourseChat, a multi-course RAG tutor deployed for a
business school. Students interact through a Moodle-embeddable chat that
supports (i)~conversation grounded in the selected offering's indexed materials
and (ii)~module-scoped practice quizzes with answer keys, hints, and source
excerpts. Six offerings (anonymized here as Courses~1--6) share one application
binary while remaining isolated at the data and routing layers.

Our central claim is that building such a tutor requires an explicit
hardware--software trade-off strategy: choose models and pipelines that
fit the host memory and concurrency envelope; push quality into retrieval,
scoping, and evaluation rather than into ever-larger generators; and keep
product surfaces (conversation vs.\ quiz) aligned with what the stack can
guarantee.

The practical question is which component should change when a tutor gives a
poor answer. An unhelpful follow-up can originate in lost conversational
context, a missing passage, a relevance filter that rejects a useful passage,
or a generator that misstates correctly retrieved evidence. A larger model
addresses only some of these failures, and its benefits must be weighed
against response time and serving compatibility. We use this distinction to
organize both the architecture and the evaluation.

The paper makes three contributions. First, it describes a replicated local
RAG architecture and its course-scoping and artifact-management boundaries.
Second, it reports a generation-model comparison with fixed retrieval and
token limits, alongside a deliberately separate model-and-response-contract
experiment. Third, it connects observed failures to concrete design choices:
retrieval-topic resolution, bounded evidence selection, and prebuilt practice
banks. Together these provide an evidence-based account of why the current
release retains an 8B model despite having enough memory to trial larger ones.

CourseChat began as a pilot project at the Dhillon School of Business to
explore how a locally developed AI tutor could add value to students' learning.
The university already provides students with access to Copilot and Gemini,
and students also use ChatGPT and other generative AI tools. The pilot's
purpose was therefore to complement these existing options with support
tailored to local courses, teaching materials, and learning expectations.
Rather than replicate general-purpose AI assistance, CourseChat was designed
to help students engage with their assigned material and develop the confidence
to work independently. Course-grounded explanations, practice questions,
hints, and feedback were intended to encourage students to attempt problems
themselves, examine their reasoning, and return to the course sources. The
intended educational value lay in supporting students' own effort and
understanding, with AI serving as a guide within that process.

\section{Related work and positioning}
\label{sec:related}

CourseChat brings together three research concerns: how tutoring supports
learning, how generated explanations remain faithful to course evidence, and
how an institution can serve those interactions reliably. The relevant
literature therefore extends beyond comparisons of language-model capability.
It also provides reasons to distinguish successful task completion, useful
instructional interaction, and learning that persists after assistance is
removed.

\subsection{Tutoring, practice, and self-regulated learning}

Research on intelligent tutoring systems predates conversational language
models and emphasizes the organization of instructional support. VanLehn's
review distinguishes systems by the granularity of their interaction, including
whether feedback addresses a final answer or intermediate problem-solving
steps~\cite{vanlehn2011}. Kulik and Fletcher's meta-analysis of fifty controlled
evaluations found generally positive learning effects for intelligent tutoring
systems, while also showing that outcomes depend on the comparison condition,
assessment, and implementation~\cite{kulik2016}. These findings support studying
the instructional process around a response rather than treating fluent
dialogue as sufficient evidence of tutoring effectiveness. They do not imply
that a language-model chatbot inherits the benefits of a validated tutor.

Practice and feedback provide a more specific rationale for the pilot's
student controls. In experiments using prose passages, Roediger and Karpicke
found that retrieval practice improved delayed retention relative to repeated
study, even when restudy produced better immediate performance~\cite{roediger2006}.
The result motivates opportunities to attempt a question before viewing its
answer, although it does not establish the effectiveness of CourseChat's
multiple-choice banks. Nicol and Macfarlane-Dick's account of formative
feedback links useful feedback to students' ability to judge their work and
regulate subsequent learning~\cite{nicol2006}. In this framing, a hint or
explanation should help a learner identify a gap and decide what to do next.
CourseChat adopts practice attempts, optional hints, and explanatory feedback
as design choices informed by this literature; whether they improve study
habits or retention in this setting remains an empirical question.

\subsection{Generative AI assistance and independent learning}

Recent experiments show why the manner of providing AI assistance matters.
Bastani et~al. studied generative AI support in high-school mathematics,
comparing a relatively unrestricted assistant with a tutor configured to provide
teacher-informed instructional support~\cite{bastani2025}. Access to AI
improved performance during assisted practice, but students using the
relatively unrestricted version subsequently performed worse than the no-AI
control group on an unassisted exam.
The tutoring safeguards mitigated that reduction; this should not be
interpreted as evidence that the guarded condition produced a positive
unassisted learning gain. The study distinguishes help with the current
exercise from developing the ability to solve another exercise independently.
That distinction is directly relevant to a pilot intended to encourage
students' own work rather than merely make answers easier to obtain.

Kestin et~al. report a complementary result from a randomized crossover study
in undergraduate physics: a deliberately designed AI tutor produced higher
immediate learning gains than the comparison active-learning class
sessions~\cite{kestin2025}. The intervention embedded pedagogical guidance
and structured content, so it was more specific than access to a general
chat interface. Its short instructional scope also limits conclusions about
long-term retention, other disciplines, or semester-long use. Read together,
these studies suggest that the instructional design and help policy are part
of the intervention being evaluated. They do not support either a universal
claim that AI tutoring improves learning or a universal claim that it
undermines it.

For the Dhillon pilot, this evidence shifts the design question toward the
additional educational value of a local service. Students already have other
ways to obtain generated text. CourseChat's intended contribution is to connect
that help to the selected course and give students a practical route between
attempting a question, requesting limited assistance, checking an explanation,
and returning to the assigned material. This is a rationale for the design,
not a claim that the pilot has reproduced either study's learning outcomes.

\subsection{Course grounding and source fidelity}

Retrieval-augmented generation provides a technical basis for connecting that
help to local materials. Lewis et~al. combined a generative model with an
external document index, demonstrating improvements on knowledge-intensive
tasks and a mechanism for updating accessible evidence separately from model
parameters~\cite{lewis2020rag}. CourseChat uses the broader retrieval-conditioned
generation pattern rather than reproducing that paper's training procedure.
Its course collections and module filters determine which teaching materials
are eligible to support a response. The resulting benefit is control over
the evidence supplied to generation; it does not make the generated answer
automatically correct.

Gao et~al.'s ALCE benchmark evaluates generated answers along separate
dimensions of correctness, fluency, and citation quality~\cite{gao2023alce}.
Their results show that even citation-producing systems can leave claims
without adequate support. This distinction matters in education because a
valid link to a textbook passage can coexist with an incorrect interpretation
of its conditions. CourseChat therefore treats source selection, citation
resolution, and fidelity of the explanation as distinct checks. Source cards
make evidence inspectable, while answer review must still establish whether
the cited passage supports the claim.

Increasing the context window is also an incomplete response to grounding
failures. Liu et~al. found that the position of relevant information affected
performance on document question answering and key-value retrieval, with
information in the middle of long contexts often used less
effectively~\cite{liu2024lost}. Their experiments concern particular models
and tasks, not the present classroom deployment. Nevertheless, they motivate
testing evidence selection and organization instead of assuming that supplying
more textbook content will improve an answer. Bounded context, explicit scope,
and a fixed retrieval configuration make CourseChat's model comparisons more
interpretable.

\subsection{Evaluation and serving constraints}

RAG evaluation frameworks reinforce the need to assess components separately.
RAGAs proposes reference-free measures for retrieved context, faithfulness,
and answer quality~\cite{es2024ragas}. ARES uses trained judges to evaluate
context relevance, answer faithfulness, and answer relevance, together with
human annotations to account for judge error~\cite{saadfalcon2024ares}.
These approaches provide useful models for diagnosis, but their scores are
not measures of student understanding. CourseChat's reported evaluation does
not claim to implement either framework. It follows the same separation of
questions: whether the intended evidence was found, whether the response was
supported, and whether the interaction was useful. Faculty assessment and
unassisted learning measures address additional questions that automated
retrieval or citation checks cannot settle.

Serving research also shows why the suitability of a model cannot be inferred
from parameter count alone. Kwon et~al.'s work on PagedAttention demonstrates
that managing the memory used by active generation requests affects batching
and throughput~\cite{kwon2023pagedattention}. Their vLLM results do not predict
the performance of CourseChat's Ollama stack or establish that this deployment
would benefit from a particular serving migration. They do establish a relevant
systems concern: model weights are only part of the resource demand during
concurrent use. CourseChat accordingly treats latency, request overlap,
context limits, and client compatibility as evaluation criteria alongside
answer quality, while acknowledging that its serial trials do not measure
classroom capacity.

\subsection{Position of the CourseChat pilot}

Related course-specific tutors combine retrieval with explicit instructional
control. CourseAssist uses course-grounded retrieval, user intent
classification, and question decomposition to support computer science
students~\cite{feng2024courseassist}. KITE combines multimodal retrieval from
course materials with an intent-aware Socratic strategy that provides hints,
guiding questions, and progressive scaffolding for algorithmic
problem-solving~\cite{jain2026kite}. CourseChat shares their emphasis on
course-aligned assistance and develops the operational side of that design:
replicated on-premises serving across segregated offerings, model-selection
trials under a fixed latency allowance, and prebuilt quiz banks with source
bindings and item-review records. Its contribution is evidence about these
deployment choices and their observed failure modes in business education.

The pilot's contribution is the integration and evaluation of these concerns
within a local business-school setting. It combines course-specific evidence,
conversation continuity, prebuilt practice, and a shared local generator
across six separately scoped offerings. The paper contributes an account of
the resulting tradeoffs and failure modes, including why a working 8B
configuration was retained despite credible alternatives. It does not propose
a new retrieval algorithm, demonstrate superiority over institutionally
available AI services, or report a controlled learning-outcome comparison.

This position also shapes the next evaluation. A useful local tutor should
be assessed for its ability to support appropriate independent effort as well
as produce accurate explanations. Measures of unaided problem solving,
retention, source use, and patterns of hint requests would connect the
engineering results reported here to the educational aims established in
the tutoring literature. The present study supplies the implementation and
diagnostic evidence needed to design that evaluation.

\section{System overview}
\label{sec:overview}

\subsection{Student experience}

The student experience is organized around the material assigned in a
particular course offering. In the intended Moodle integration, students enter
through an instructor-provided link embedded in the LMS. CourseChat is meant to
sit inside that Moodle course context rather than as a public textbook or
tutoring destination: the embed is the intended entry path, and the course slug
is not advertised to students as a standalone bookmark, although slug secrecy
and enrollment checks are not claimed as proven access controls. That link
establishes the full-course, module, or named-activity scope; the header
displays the course, instructor, course reference number (CRN), and selected
material. This
visible context helps students understand which materials the tutor can use
and connects the interaction to their immediate coursework. Conversation and
quiz functions are implemented in the course pages, while complete acceptance
of the public gateway remains a separate deployment task. The design seeks
local value through this connection to assigned materials and practice,
complementing students' access to general-purpose AI services.

In Conversation mode, students ask questions in ordinary language and receive
explanations accompanied by citations to passages from the selected offering. Inline
citations open source cards containing excerpts and available chapter,
section, or page information. Keeping these cards collapsed initially gives
the explanation room while preserving a route back to the original material.
Students can request a shorter answer, more detail, or another explanation,
and can adjust the explanation level. These controls are intended to support
checking and revisiting a concept rather than treating the first generated
answer as the endpoint of study. The interface also distinguishes source
relevance from answer correctness: its relevance score describes the match
between a passage and a query, not the probability that an explanation is
correct. Course-configured relevance floors set a trusted lower bound;
students may raise but not lower that floor.

Conversational continuity matters because students need not restate a complete
question at every turn. After an explanation of net present value, for example,
``Can you show me that with numbers?'' should preserve that topic and the
selected module. A brief social aside should not erase it. The router uses
recent dialogue to resolve the reference, then retrieves current course
evidence; earlier assistant text does not become an authoritative source.
Where several referents are plausible, it asks a short clarification question.
Social replies do not acquire artificial course citations. When retrieval
finds no qualifying passage within the selected module, a reported fallback
may broaden the explanation using the model's current conversational context
and material from the same course offering. It does not retrieve material
or conversation history from another segregated course. If qualifying course
evidence remains unavailable, the system reports the evidence gap. The
presence of a qualifying passage cannot by itself guarantee a faithful
explanation.

Quiz mode provides a different form of participation. Selecting Quiz starts a
practice session from a prepared bank and presents one multiple-choice item
at a time. The student can attempt the item, request a stored hint, or skip it.
A hint leaves the current question and progress unchanged and does not reveal
the answer key or supporting excerpt. After an answer, the system reports
whether it is correct, supplies an explanation and source evidence, and
presents the next item; skipping also reveals the explanation. This sequence
is intended to make independent attempts and reflection convenient, although
it does not require an attempt before disclosure. The 435 prepared items cover
65 configured modules, and practice uses stored keys and feedback without live
generation. The banks are assistant-curated and await instructor review;
their technical audits establish reproducible behavior rather than educational
validity.

Students can move from practice to discussion while retaining the conversation
and instructor-selected scope, allowing a difficult item to prompt a broader
explanation of the concept. Selecting Quiz again starts a new session.
Clearing the conversation retains the scope and mode, while history is
separated by course and scope. This supports a sequence of asking, attempting,
checking, and revisiting within a familiar course context. Its value for
motivation, independent effort, or learning remains an intended outcome to be
evaluated with students, rather than a measured result.

The interface also provides progress messages and incremental answer text
through server-sent events. Visible generation can signal that a request is
being processed before the complete explanation is available. This perceived
responsiveness depends on the application and gateway forwarding events
incrementally; server-local complete-answer timings do not establish the
student's time to first visible text through Moodle.

\subsection{Logical modules}

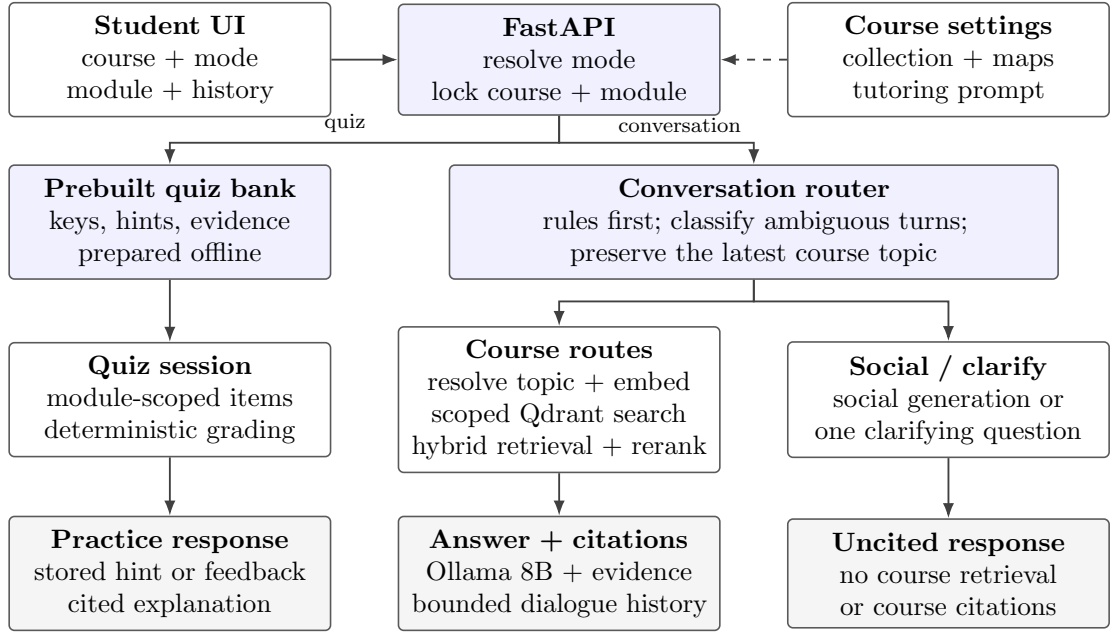
\begin{figure}[t]
\centering
\begin{tikzpicture}[
  font=\small,
  box/.style={draw=black!65, line width=0.6pt, rounded corners=2pt,
    align=center, text width=3.9cm, minimum height=1.3cm, inner sep=5pt,
    fill=white},
  control/.style={box, fill=blue!6},
  result/.style={box, fill=gray!8},
  flow/.style={-{Latex[length=2mm]}, line width=0.7pt, draw=black!75},
  dependency/.style={flow, dashed}
]
  \node[box] (ui) at (0,0)
    {\textbf{Student UI}\\course + mode\\module + history};
  \node[control] (api) at (5.15,0)
    {\textbf{FastAPI}\\resolve mode\\lock course + module};
  \node[box] (config) at (10.3,0)
    {\textbf{Course settings}\\collection + maps\\tutoring prompt};

  \node[control] (quiz) at (0,-2.15)
    {\textbf{Prebuilt quiz bank}\\keys, hints, evidence\\prepared offline};
  \node[control, text width=7.7cm] (router) at (7.725,-2.15)
    {\textbf{Conversation router}\\rules first; classify ambiguous turns;\\preserve the latest course topic};

  \node[box] (session) at (0,-4.5)
    {\textbf{Quiz session}\\module-scoped items\\deterministic grading};
  \node[box] (retrieve) at (5.15,-4.5)
    {\textbf{Course routes}\\resolve topic + embed\\scoped Qdrant search\\hybrid retrieval + rerank};
  \node[box] (casual) at (10.3,-4.5)
    {\textbf{Social / clarify}\\social generation or\\one clarifying question};

  \node[result] (practice) at (0,-6.8)
    {\textbf{Practice response}\\stored hint or feedback\\cited explanation};
  \node[result] (answer) at (5.15,-6.8)
    {\textbf{Answer + citations}\\Ollama 8B + evidence\\bounded dialogue history};
  \node[result] (uncited) at (10.3,-6.8)
    {\textbf{Uncited response}\\no course retrieval\\or course citations};

  \draw[flow] (ui) -- (api);
  \draw[dependency] (config) -- (api);
  \draw[flow] (api.south) -- (5.15,-1.1) -- (0,-1.1)
    node[pos=0.55, above, font=\scriptsize]{quiz} -- (quiz.north);
  \draw[flow] (api.south) -- (5.15,-1.1) -- (7.725,-1.1)
    node[pos=0.62, above, font=\scriptsize]{conversation} -- (router.north);
  \draw[flow] (quiz) -- (session);
  \draw[flow] (session) -- (practice);
  \draw[flow] (router.south) -- (7.725,-3.2) -| (retrieve.north);
  \draw[flow] (router.south) -- (7.725,-3.2) -| (casual.north);
  \draw[flow] (retrieve) -- (answer);
  \draw[flow] (casual) -- (uncited);
\end{tikzpicture}
\caption{Application request paths. Quiz practice uses stored items and
deterministic grading. Course, mixed, and dependent follow-up routes retrieve
scoped evidence before generation; social and clarification routes do not
retrieve course content. Responses return through the same API and UI.
The dashed arrow supplies course configuration.}
\label{fig:modules}
\end{figure}

Figure~\ref{fig:modules} shows how request handling, retrieval, generation,
and course artifacts compose.

\subsection{Deployment topology}

\begin{figure}[t]
\centering
\begin{tikzpicture}[
  font=\small,
  box/.style={draw=black!65, line width=0.6pt, rounded corners=2pt,
    align=center, inner sep=6pt, fill=white},
  traffic/.style={{Latex[length=2mm]}-{Latex[length=2mm]},
    line width=0.7pt, draw=black!75},
  transfer/.style={-{Latex[length=2mm]}, dashed,
    line width=0.7pt, draw=black!75}
]
  \node[box, text width=4.8cm] (browser) at (0,0)
    {\textbf{Browser / Moodle embed}\\course-specific public URL};
  \node[box, fill=gray!8, text width=6.4cm] (iis) at (0,-1.8)
    {\textbf{Campus IIS gateway}\\TLS termination, course-slug rewrites,\\ARR load balancing and health checks};

  \node[box, fill=blue!6, text width=5.5cm, minimum height=2.8cm]
    (a) at (-3.65,-4.55)
    {\textbf{Node A: serve + ingest}\\[3pt]
     FastAPI + static course UI\\
     Local embeddings and reranker\\
     Qdrant: one collection per offering\\
     Ollama: \texttt{llama3.1:8b}\\[3pt]
     Canonical corpus ingestion};
  \node[box, fill=blue!6, text width=5.5cm, minimum height=2.8cm]
    (b) at (3.65,-4.55)
    {\textbf{Node B: serve + replicate}\\[3pt]
     FastAPI + static course UI\\
     Local embeddings and reranker\\
     Qdrant: one collection per offering\\
     Ollama: \texttt{llama3.1:8b}\\[3pt]
     Restored course indexes};

  \draw[traffic] (browser) -- node[right, font=\scriptsize]{HTTPS} (iis);
  \draw[traffic] (iis.south) -- (0,-2.8) -| (a.north);
  \draw[traffic] (iis.south) -- (0,-2.8) -| (b.north);
  \node[fill=white, inner sep=2pt, font=\scriptsize] at (0,-2.8)
    {internal HTTP / responses};

  \draw[transfer] (a.south) -- (-3.65,-6.55) -- (3.65,-6.55)
    node[midway, below, font=\small]{Collection snapshots: Node A $\rightarrow$ Node B}
    -- (b.south);
  \node[align=center, text width=13cm, font=\small]
    at (0,-7.45)
    {Quiz banks, source ledgers, review records, and application assets\\
     are deployed separately to both nodes.};
\end{tikzpicture}
\caption{Campus access architecture and deployed twin local stacks. Solid
arrows indicate request/response traffic; the dashed path transfers index
snapshots. Quiz and application artifacts are deployed separately. The node
audits reported here bypass the gateway; complete public course-slug
acceptance remained open in the deployment snapshot.}
\label{fig:topo}
\end{figure}
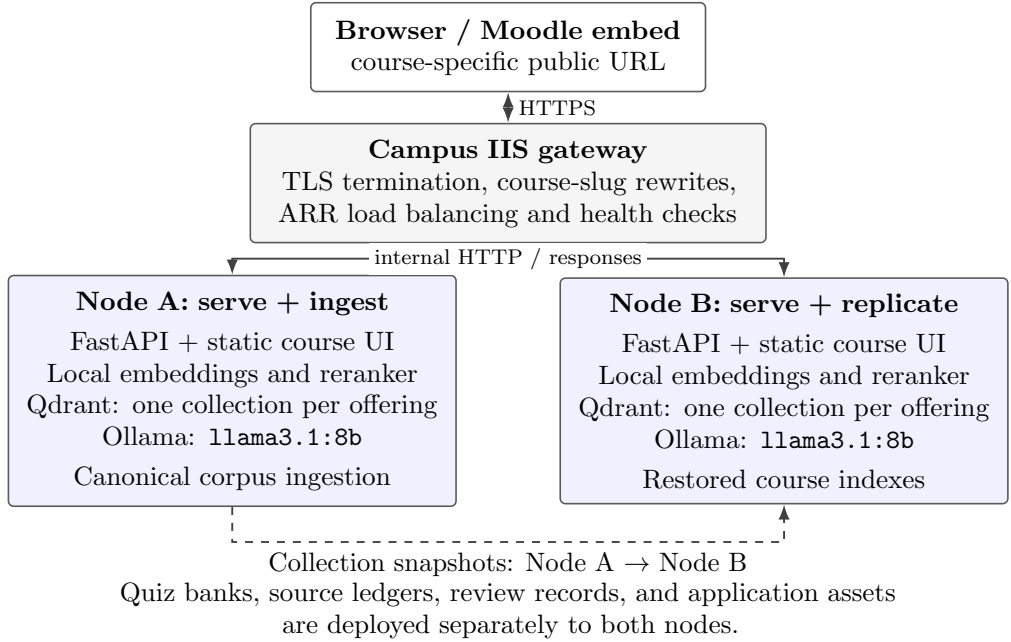

Figure~\ref{fig:topo} summarizes the campus deployment architecture. The IIS
gateway is assigned TLS termination, course-slug rewrites, and routing to
two AI nodes (Node~A and Node~B). Each node runs the full stack locally:
FastAPI, Qdrant, sentence-transformer embeddings, a cross-encoder reranker, and
Ollama. Backend
health and server-local audits establish the state of the two nodes; they do
not establish that every public slug or proxy route has passed acceptance.

\textbf{Hardware envelope (classroom hosts).}
 Each node is a DGX Spark-class
appliance with a Grace Blackwell GB10 and on the order of 128\,GB unified
memory shared by the API workers, vector database, embedding/rerank models, and
Ollama. Two uvicorn workers allow overlapping application requests, although
actual generation overlap depends on the inference server's scheduling and
memory use. These are independent local serving stacks, not a pooled GPU
model spread across both machines. This envelope frames the model trials in
Section~\ref{sec:bakeoff}; worker count alone does not establish classroom
capacity.

\section{Hardware--software trade-off strategy}
\label{sec:tradeoffs}

We treat model selection as a constrained operating decision. The desired
outcome is a useful explanation supported by the selected offering's materials,
delivered within a tolerable classroom wait. The bake-off makes one performance
constraint explicit: median complete-answer latency must be no more than
30\% above the production 8B baseline measured in the same round. Other
promotion criteria require usable overlapping requests, preservation of course
scope, and a meaningful improvement in reviewed answers.
These are promotion requirements; a listed requirement is not evidence that
the corresponding experiment was completed.

Several design choices act on different parts of this objective. Model family
and size affect answer generation, whereas retrieval depth, context limits,
and reranking affect which evidence reaches the model and how soon generation
can begin. Server-wide application profiles package these performance settings;
they do not select a course or change the collection boundary. A separate
product decision moves quiz authoring out of the request path. A separate
deployment decision moves document extraction and review to a workstation,
leaving the serving hosts responsible for indexing and inference.

Changing these components together would obscure why a candidate succeeded or
failed. We therefore keep the serving retrieval stack fixed during the model
bake-off, and identify the separate structured-answer experiment as a different
comparison. Course content remains local, quiz questions are prepared before
practice, and reasoning-capable models must deliver student-visible output
within the configured budget before they are considered deployable. These
choices trade some flexibility for traceable failures and a bounded release
process.

Table~\ref{tab:tradeoff-summary} brings together the main interventions,
their observed outcomes, and the resulting design decisions. It distinguishes
generator comparisons from changes to retrieval and practice delivery;
Section~\ref{sec:bakeoff} gives the model-by-model measurements. This overview
makes explicit why the project retained some components, changed others, and
deferred alternatives that had not established an overall advantage.

\begin{table}[tbp]
\centering
\small
\setlength{\tabcolsep}{4pt}
\renewcommand{\arraystretch}{1.18}
\caption{Decision summary across model trials and system-design evaluations.
Results come from distinct protocols and are not a common accuracy ranking.
Model latency ratios use the 8B baseline from the corresponding bake-off
round; all reported timings are server-local.}
\label{tab:tradeoff-summary}
\begin{tabular}{@{}p{0.17\linewidth}p{0.28\linewidth}p{0.31\linewidth}p{0.16\linewidth}@{}}
\toprule
\raggedright Approach &
\raggedright Observed result &
\raggedright Trade-off &
\raggedright Decision \tabularnewline
\midrule
\raggedright Larger dense generators\newline(first round) &
\raggedright All passed 12/12 live probes; median completion times were
1.75--3.36 times the 8B baseline. &
\raggedright Some answers were more complete, but every candidate exceeded
the $+30\%$ classroom latency limit. Probe passes alone do not establish
pedagogical quality. &
\raggedright Retain 8B for classroom service. \tabularnewline
\addlinespace
\raggedright Qwen 7B and Mistral Nemo 12B &
\raggedright Both passed 12/12 probes; latency ratios were 1.01 and 1.24.
Qwen completeness was 0.93 versus 0.91 for 8B. &
\raggedright Both met the speed gate. Speed passes and small completeness
differences did not establish a faculty-reviewed tutoring advantage. &
\raggedright Retain as candidates for paired review. \tabularnewline
\addlinespace
\raggedright Structured Qwen mixture-of-experts response &
\raggedright Warm median was 3.00\,s versus 3.47\,s for 8B, but review found
three regressions across twelve fixed-evidence cases. &
\raggedright Faster warm responses and better corrections accompanied a new
factual error and an unnecessary refusal. Model and response contract changed
together. &
\raggedright Hold activation. \tabularnewline
\addlinespace
\raggedright Deterministic follow-up resolution &
\raggedright Topic checks rose from 23.3--40.0\% in baseline resolver
preflight to 90--100\% live across five courses; all 400 scope checks passed. &
\raggedright Better continuity required no generator change. Correct topic
selection and scope do not by themselves verify answer fidelity. &
\raggedright Deploy resolver with 8B. \tabularnewline
\addlinespace
\raggedright Prepared quiz banks &
\raggedright 435 questions cover 65 modules. Warm practice-request medians
were 4.8 and 5.8\,ms on the two nodes. &
\raggedright Inspectable evidence and deterministic grading require finite
banks, maintenance after source changes, and instructor review. &
\raggedright Serve prepared banks; faculty approval remains pending.
\tabularnewline
\bottomrule
\end{tabular}
\end{table}

\section{Design choices}
\label{sec:design}

\subsection{Isolation without six fine-tuned LLMs}

Isolation is enforced by \emph{separate course ids}, \emph{separate on-disk
trees}, \emph{separate Qdrant collections}, and \texttt{?course=} routing. A
shared Ollama generator receives passages from the routed collection.
Optional per-course embedding fine-tunes are a potential retrieval-quality
intervention; they are not required for isolation and were not changed in the
model trials. The collection and request boundaries determine which indexed
materials can be selected, while source-faithful generation remains a
separate requirement.

Each uploaded offering is keyed by its own course reference number (CRN).
Instructors can therefore maintain different versions of the same catalog
course---for example, distinct sections or successive terms---and control which
materials enter that offering's retrieval collection. Uploading documents
updates the evidence available to that CRN; it does not retrain the shared
language model or merge corpora across offerings. Two offerings may share a
catalog code (e.g., two sections of the same strategic management course) yet
must keep distinct corpora when instructors and materials differ. The same
principle applies to conversation history and quiz state: carrying a previous
course's session into a newly selected offering would defeat collection-level
separation. Scope therefore belongs to request handling and session validation
as well as storage. Passing the reported scope checks supports these specific
paths, rather than proving that a shared pretrained model can never produce an
unsupported statement.

\subsection{RAG for content; prompts for pedagogy}

Course knowledge is indexed, not baked into LLM weights. Tutoring tone and
Socratic instructions come from system prompts and optional bot-profile YAML.
Chapter scope is applied through request validation and retrieval metadata
filters; any reported fallback remains within the selected course offering.
Content updates are re-ingest + replicate, not a training
job. This allows revisions to textbooks and decks to be associated with explicit
source artifacts and ingestion receipts. A weight update would not by itself
provide the same passage-level provenance, and would add another artifact to
version across the two serving nodes.

\subsection{Prebuilt quizzes}

Early demos showed that generating a multiple-choice item at chat time is
insufficient for a reliable practice session. The platform therefore
prebuilds banks containing source-backed
items, keys, explanations, hints, and evidence ledgers. Automatic generation
is gated by index and answer evaluation. A separate publication path accepts
independently reviewed items bound to the exact current source and question
hashes; that path requires the author and reviewer identities to differ, and
that item-level review does not clear a failed free-form chat audit.
Serving quiz turns is
deterministic grading against the bank (warm medians on the order of
milliseconds server-local), while conversation retains the generative path.
Answer keys remain server-side until an attempt or reveal, so the bank is not
exposed as a downloadable key file in the student UI.

Across Courses~1--6 the deployed banks total 435 questions over 65
modules. Structural audits and live API grading checks pass on both nodes;
instructor pedagogical review remains an explicit separate gate
(\texttt{instructor\_reviewed: false} until faculty sign-off).

This choice makes practice behavior reproducible: the hint, key, explanation,
and supporting excerpt can all be inspected before use. It also makes content
maintenance explicit. When source material changes, a bank must be checked
against the new corpus rather than silently continuing to cite stale evidence.
The cost is a finite pool of questions and less spontaneous adaptation to a
student's wording, which is why open-ended explanation remains a separate
conversation mode.

\subsection{Conversational turns over shared course evidence}
\label{sec:conversation-design}

Conversation gives students a way to work through the meaning of course
material, including questions that cannot be anticipated in a prepared quiz.
The design combines three resources with different roles: a pretrained model
provides language generation, indexed passages provide current teaching
evidence, and prebuilt quiz items provide stable questions, hints, and grading
keys. Ingesting a professor's materials updates the evidence collection; it
does not retrain the language model. Conversation can therefore extend the
uses of the same corpus while preserving the distinction between an open-ended
explanation and a stored practice item.

A conversational turn carries more than a new search phrase. If a student
asks whether a different assumption would change a worked example, the tutor
must retain the example's setting, identify the changed assumption, and find
the course evidence needed to explain its effect. An exchange about net
present value might move from a definition to a numerical illustration and
then to the student's own account of why a higher discount rate changes the
result. This illustrates the intended interaction: the student can test an
interpretation and request a targeted explanation. The model's hypothetical
numbers must remain clearly distinguished from facts attributed to course
sources, and a change of topic must still respect the selected module.

The implemented path resolves dependent turns against recent dialogue while
passing the student's actual wording to generation. Recent dialogue is held in
the browser and submitted with each request rather than stored as a durable
server session: history must match the selected course, is bounded in length,
and does not require sticky affinity between the twin nodes. The shared
resolver raised intended-topic follow-up rates to 90--100\% across the
five-course replay, without changing the generator (Section~\ref{sec:eval}).
The conversation router handles social, course, mixed, follow-up, and
clarification turns, using deterministic rules first and a short classification
call for ambiguity. A casual aside can preserve the course topic; subsequent
content questions retrieve fresh evidence. Earlier assistant text helps
interpret a reference but is not promoted into authoritative teaching material.
Only designated meta or policy FAQ entries short-circuit retrieval; content
FAQs remain in the course vector index and compete with other evidence.

LangChain components already provide model, prompt, document, and vector-store
interfaces in the retrieval application. They offer a basis for extending the
conversation workflow around the existing corpus: assemble the learning
instructions and bounded dialogue, retrieve material for the resolved turn,
and generate an explanation with source references. LangChain's memory
facilities support thread-level history and strategies for limiting its size;
durable conversation state would additionally require shared persistence and
course isolation~\cite{langchainmemory}. The current router uses explicit
Python control flow and browser-supplied history. A more extensive LangChain
workflow, or LangGraph orchestration for resumable multi-step activities,
would be a future implementation choice, evaluated against this simpler
baseline.

This separation creates room for additional instructional modes without
building another knowledge base for each one. A future case-discussion mode
could ask students to justify an interpretation using a professor's
supplementary case; a guided-practice mode could focus on the next reasoning
step; and a revision mode could revisit a misconception exposed by a quiz.
Each would use an explicit interaction policy over the same scoped evidence
and keep factual support checks in place. Their local value would come from
the instructor's selected material and teaching expectations, combined with
opportunities for student reasoning. These are proposed extensions to the
deployed conversation and quiz modes, to be tested for usefulness and
appropriate independent effort.

\subsection{Streaming vs.\ raw latency}

Non-streaming live eval reports full-answer wall times (often $\sim$6--10\,s on
8B). Perceived speed is dominated by time-to-first-token through the gateway.
The UI requests SSE, and gateway configuration must preserve incremental
delivery. The first metadata event, the first visible text, and completed
answer time are different measurements. The bake-off below measures the last
of these on the node itself; it does not quantify browser-perceived waiting
time through IIS. Streaming can expose an answer earlier without reducing
total inference work, so it complements rather than replaces model-latency
evaluation.

\subsection{Workstation / Node A / Node B pipeline}

Scanned textbook OCR and page review run on a Mac workstation (Docling, Apple
Vision, optional on-device formula assistance). Reviewed corpora are prepared
offline, ingested on Node~A with the same encoder the API uses, then
snapshot-replicated to Node~B. This split keeps OCR engines off the classroom
hosts and keeps A/B parity enforceable (\texttt{parity-check} on git SHA and
health).

\section{Model bake-off}
\label{sec:bakeoff}

\subsection{Protocol}

On Node~A we compared Ollama instruct tags while holding fixed:
\texttt{APP\_PROFILE}, context length 8192, output cap 768 tokens, retrieval
knobs, and the same live-eval probe set. Node~B remained on the production 8B
model as failover. The embedding encoder and reranker were held fixed to avoid
confounding the generator comparison; changing the encoder would also require
rebuilding stored vectors. Thinking-mode defaults
were treated as disqualifying when they emptied student-visible content or
exhausted the token budget.

Promotion rule: a candidate replaces 8B only if it is clearly better at
grounded tutoring \emph{and} clears the classroom speed gate (p50 total latency
no worse than $+30\%$ vs.\ 8B) with usable two-request overlap.

Requests were sent to the loopback API on Node~A, bypassing IIS. The twelve
live-evaluation probes exercise keyword presence and citation behavior, not a
complete pedagogical rubric. We report each candidate's median complete-answer
wall time divided by the 8B median from the same round. A fresh 8B baseline
was measured in each round, so ratios use the corresponding round rather than
a single pooled reference. Context and output
caps were fixed at 8,192 and 768 tokens, respectively; fixed caps do not imply
identical realized output lengths.

The protocol calls for reviewing directness, groundedness, synthesis, and
course-appropriate tone, alongside scope and overlap checks. The documented
quality target includes at least 95\% supported factual claims and a clear
paired-review improvement. These targets were not established by the twelve
automated passes. The reported tables contain point estimates without repeated
run uncertainty intervals or a completed concurrency study, so a speed pass
is only one prerequisite for promotion.

\subsection{Results}

Tables~\ref{tab:bakeoff1}--\ref{tab:bakeoff2} summarize two rounds on Node~A
(18--19 September 2026). Live eval scores are keyword/citation smoke checks
(necessary but not sufficient).

\begin{table}[t]
\centering
\caption{Bake-off round 1 (includes thinking-capable tags).}
\label{tab:bakeoff1}
\begin{tabular}{lcccc}
\toprule
Model & Live eval & p50 (ms) & vs.\ 8B & Speed gate \\
\midrule
llama3.1:8b (prod.) & 12/12 & 8392 & 1.00$\times$ & pass \\
qwen2.5:14b-instruct & 12/12 & 14726 & 1.75$\times$ & fail \\
gemma3:12b & 12/12 & 18855 & 2.25$\times$ & fail \\
qwen3:14b & 12/12 & 28194 & 3.36$\times$ & fail \\
qwen3.6:35b-A3B (Q4) & 0/12 & 15109 & 1.80$\times$ & fail$^\dagger$ \\
\bottomrule
\end{tabular}

\vspace{0.3em}
{\footnotesize $^\dagger$Thinking filled the reply; empty student-facing content under the live client.}
\end{table}

\begin{table}[t]
\centering
\caption{Bake-off round 2 (non-thinking / think-disabled candidates).}
\label{tab:bakeoff2}
\begin{tabular}{lcccc}
\toprule
Model & Live eval & p50 (ms) & vs.\ 8B & Speed gate \\
\midrule
llama3.1:8b & 12/12 & 8556 & 1.00$\times$ & pass \\
qwen2.5:7b-instruct & 12/12 & 8668 & 1.01$\times$ & pass \\
mistral-nemo:12b & 12/12 & 10589 & 1.24$\times$ & pass \\
ibm/granite3.3:8b & 0/12 & 21443 & 2.51$\times$ & fail$^\ddagger$ \\
\bottomrule
\end{tabular}

\vspace{0.3em}
{\footnotesize $^\ddagger$Integration mismatch with the chat client (empty \texttt{content}), not a claim the model cannot answer via raw Ollama.}
\end{table}

\paragraph{Deployment decision.} Production remains \texttt{llama3.1:8b} on both nodes. No
round-1 larger model cleared the classroom speed gate. Round~2 identified
\texttt{qwen2.5:7b-instruct} as a same-speed side-grade with a small
completeness bump, and \texttt{mistral-nemo:12b} as a larger option still inside
$+30\%$. Promotion waits on faculty side-by-side review of answer quality.

Round~1 illustrates the difference between memory feasibility and acceptable
response time: Qwen2.5~14B and Gemma~12B returned all twelve expected smoke-test
responses but required approximately 1.75 and 2.25 times the baseline median.
Round~2 provides a counterexample to rejecting larger models categorically:
Nemo~12B completed at 1.24 times its fresh baseline. Qwen~7B completed at
1.01 times the baseline and had a modest reported completeness-proxy increase
(0.93 versus 0.91). That proxy is not a supported-claim rate or an instructor
rating. The result is a shortlist for further review, rather than evidence
that parameter count alone determines suitability.

The zero-pass rows require a different interpretation. The tested Qwen MoE
configuration spent its budget in a thinking channel and supplied no
student-visible answer through the live client. Granite also returned empty
content through the application despite answering through raw Ollama. These
are failures of the evaluated serving configuration. They matter to deployment,
but they do not establish that the models cannot answer the questions.

\subsection{Separate fixed-evidence fidelity comparison}
\label{sec:fidelity}

A separate diagnostic compared the production 8B path with a
\texttt{qwen3.6:35b-a3b-q4\_K\_M} candidate on twelve frozen cases. The
fixture preserved the current question, earlier dialogue, selected source
passages, and source checksums. It included corrections after an incorrect
assistant answer, numerical illustrations, missing facts, topic switching,
format constraints, and references to a previous example. Unlike the
generator-only bake-off, the final candidate changed the model, prompt,
structured response contract, validation, and transport together. Its results
therefore evaluate that complete configuration, not an isolated model-size
effect.

The candidate returned an evidence list of source identifiers and exact
quotations followed by an answer. Validation checked structure, current source
identity, quotation text, and cited source numbers; rejected drafts were
replaced by a fixed conversational gap response. Thinking was explicitly
disabled. Direct requests to the generation endpoint produced twelve valid
JSON objects, resolving a format-handling problem observed with the installed
chat route. Nine answers passed structural checks, including one materially
wrong claim. Three were withheld: two appropriately lacked evidence, whereas
one was an answerable follow-up.

The independent comparison recorded three substantial improvements, two
modest improvements, three ties, one mixed result, and three regressions.
The larger candidate corrected earlier accounting and pricing mistakes, but
also turned a qualified benefit of holding cash into a guarantee and lost
continuity on a reference to a prior example. Warm serial medians were
3.469\,s for the production path and 2.997\,s for the candidate. A separate
cold candidate request took 25.417\,s, including 23.269\,s loading. With only
twelve warm observations, the reported nearest-rank p95 is the maximum,
not a reliable estimate of classroom tail latency. The candidate occupied
approximately 26.4\,GB on the GPU with the configured context, demonstrating
that memory fit was not the decisive obstacle in this trial.

The configuration was not activated. Its warm speed and correction gains
were promising, but new factual and continuity regressions prevented an
overall quality win. Production streaming, dynamic instructor instructions,
and the broader interaction settings were also outside this diagnostic's
validated contract. Section~\ref{sec:discussion} uses these results to explain
why retaining 8B is a conditional release decision.

\section{Content and RAG pipeline}
\label{sec:pipeline}

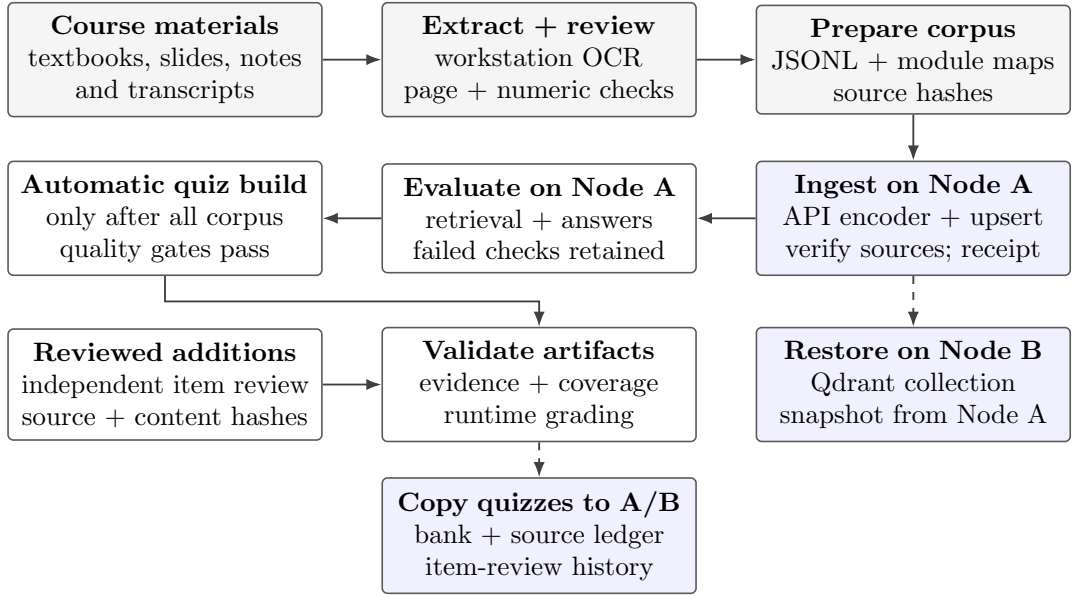
\begin{figure}[t]
\centering
\begin{tikzpicture}[
  font=\small,
  box/.style={draw=black!65, line width=0.6pt, rounded corners=2pt,
    align=center, text width=3.8cm, minimum height=1.15cm,
    inner sep=5pt, fill=white},
  prepared/.style={box, fill=gray!8},
  serving/.style={box, fill=blue!6},
  flow/.style={-{Latex[length=2mm]}, line width=0.7pt, draw=black!75},
  transfer/.style={flow, dashed}
]
  \node[prepared] (raw) at (0,0)
    {\textbf{Course materials}\\textbooks, slides, notes\\and transcripts};
  \node[prepared] (review) at (4.95,0)
    {\textbf{Extract + review}\\workstation OCR\\page + numeric checks};
  \node[prepared] (corpus) at (9.9,0)
    {\textbf{Prepare corpus}\\JSONL + module maps\\source hashes};

  \node[box] (auto) at (0,-2.1)
    {\textbf{Automatic quiz build}\\only after all corpus\\quality gates pass};
  \node[box] (eval) at (4.95,-2.1)
    {\textbf{Evaluate on Node A}\\retrieval + answers\\failed checks retained};
  \node[serving] (ingest) at (9.9,-2.1)
    {\textbf{Ingest on Node A}\\API encoder + upsert\\verify sources; receipt};

  \node[box] (reviewed) at (0,-4.3)
    {\textbf{Reviewed additions}\\independent item review\\source + content hashes};
  \node[box] (bank) at (4.95,-4.3)
    {\textbf{Validate artifacts}\\evidence + coverage\\runtime grading};
  \node[serving] (replica) at (9.9,-4.3)
    {\textbf{Restore on Node B}\\Qdrant collection\\snapshot from Node A};

  \node[serving] (deploy) at (4.95,-6.3)
    {\textbf{Copy quizzes to A/B}\\bank + source ledger\\item-review history};

  \draw[flow] (raw) -- (review);
  \draw[flow] (review) -- (corpus);
  \draw[flow] (corpus) -- (ingest);
  \draw[flow] (ingest) -- (eval);
  \draw[flow] (eval) -- (auto);
  \draw[flow] (auto.south) -- (0,-3.2) -| (bank.north);
  \draw[flow] (reviewed) -- (bank);
  \draw[transfer] (ingest) -- (replica);
  \draw[transfer] (bank) -- (deploy);
\end{tikzpicture}
\caption{Knowledge preparation and deployment. Automatic quiz building
requires passing corpus quality gates. Independently reviewed additions
provide a separate path, bound to current primary evidence, without clearing
failed chat evaluations. Both paths require artifact validation. Dashed arrows
denote deployment transfers; instructor pedagogical review remains separate
and pending for the published banks.}
\label{fig:pipeline}
\end{figure}

Production chat uses a Fast RAG path: embed query, hybrid search over the
course collection, metadata filters (module/chapter), cross-encoder rerank,
optional cache, then Ollama generation with bounded history and citations.
Relevance floors are course-configured; students may raise but not lower the
server floor. The displayed relevance score summarizes passage--query match
from the reranker; it is not a confidence that the generated explanation is
correct. Retrieval also applies evidence tiers: original lecture and reading
passages are preferred over generated study aids such as cards, and navigation
or placeholder entries are excluded from grounding. When an instructor locks a
chapter or module and the filtered set is empty, a counted fallback may widen
scope using the model's current conversational context and material from the
same course offering, and report that it did so. This fallback never retrieves
materials or dialogue from another segregated course; the course collection
boundary remains unchanged. Lecture slides enter the index through explicit lecture-to-module
maps rather than filename heuristics; in the reported preparation, eighty
lecture files contributed 2,242 passages under those maps.

The four newly completed textbook-course corpora contain 15,730
prepared passages. This is a prepared-corpus count, not the total number of
Qdrant points across all six offerings; live collections can also contain
legacy material. Course metadata distinguishes a teaching module from a
textbook chapter, so a shared chapter number does not implicitly determine
the scope of a teaching activity.

The pipeline preserves original-file hashes, reviewed page metadata, stable
chunk identifiers, and ingestion receipts. These artifacts make it possible
to check that a stored quotation still belongs to the current source before
publishing a question. Scanned tables and formulas need additional numerical
review: extracting a page does not certify its arithmetic or layout. Generated
study aids are not treated as primary evidence for the reviewed quiz items.
This provenance work adds preparation effort, but allows content defects to
be distinguished from retrieval and generation defects.

The split between automatic and independently reviewed quiz publication is
material to interpreting the deployment. None of the four textbook courses
passed the complete automatic regeneration gate in the reported ingestion
run. Their 45 added items followed the separate primary-source authoring and
independent item-review path, with distinct author and reviewer identities.
That path verified current evidence and item integrity without clearing the
failed chat evaluations or implying faculty approval. Snapshot replication
moves the index, while banks, source ledgers,
and cumulative review histories require separate deployment to both
nodes.

\section{Evaluation}
\label{sec:eval}

\subsection{Multi-turn conversation}

Frozen fixtures contain 20 dialogues and 80 turns per course, including 60
follow-ups. They cover numerical examples, misconceptions, topic changes,
brevity requests, locked-module boundaries, and transitions between quiz and
conversation. The harness records the resolved retrieval query, scope,
selected passages, model-visible context, reply, and latency. Topic-hit scores
measure whether a dependent question retains its intended subject; they do
not measure whether every resulting explanation is correct.

Course~1 was evaluated separately, and Courses~2--6 were replayed serially on
Node~B with the production 8B generator. Table~\ref{tab:followups} reports the
documented before/after topic checks. Course~1 has two live replays; the
Courses~2--6 baseline scores come from a resolver-only preflight with the
prior production resolver, followed by live evaluation of the update.
In the five-course run, all 400 scope
checks passed with zero API errors, and each course reached the 90\%
follow-up target. Median per-turn latency ranged from 3.24 to 6.19\,s. The
separate Course~1 replay improved from 31 to 60 successful follow-ups, while
its median increased from 4.92 to 5.81\,s. This is consistent with treating
quality and speed as distinct outcomes rather than assuming that every
resolver improvement also reduces latency.

\begin{table}[t]
\centering
\caption{Intended-topic follow-up checks. Course~1 has a live baseline;
Courses~2--6 have resolver-only baselines. Updated results are live for all
courses with the unchanged 8B generator. Scope checks also include initial turns.}
\label{tab:followups}
\begin{tabular}{lrrrr}
\toprule
Offering & Baseline & Updated live & Updated rate & Scope \\
\midrule
Course 1 & 31/60 & 60/60 & 100.0\% & 80/80 \\
Course 2 & 24/60 & 54/60 & 90.0\% & 80/80 \\
Course 3 & 22/60 & 55/60 & 91.7\% & 80/80 \\
Course 4 & 17/60 & 58/60 & 96.7\% & 80/80 \\
Course 5 & 14/60 & 59/60 & 98.3\% & 80/80 \\
Course 6 & 19/60 & 60/60 & 100.0\% & 80/80 \\
\bottomrule
\end{tabular}
\end{table}

The five-course run still contained 93 \texttt{no\_match} turns. Twelve were
intentional locked-module refusals, fifteen were quiz follow-ups whose wording
did not retrieve a qualifying passage, and 66 involved other requests spanning
24 topics. Cleaner concept probes found evidence for nineteen of those
topics; five still required content or index investigation. A bounded facet
expansion recovered eleven of the 24 original questions without lowering the
relevance floor. These diagnostics identify recoverable retrieval failures;
they are not an overall answer-accuracy score. Faculty ratings remain pending,
so neither the proposed 4/5 usefulness target nor the 95\% supported-claim
target is claimed as achieved.

\subsection{Source fidelity and routing checks}

A targeted, agent-reviewed ten-request live textbook audit gives a complementary view of
quality. It found three supported replies, four retrieval misses despite
available evidence, two source contradictions, and one partly grounded answer.
Every returned citation resolved to its source. Thus, citation integrity
coexisted with substantive error: locating a real passage did not ensure that
the answer preserved its meaning. The requests were selected diagnostics,
not a representative sample from which to estimate a population error
rate.

The 20 September router extension has a different evaluation scope. Its
frozen contract contains 32 crossover cases, supplemented by focused checks
and live probes for social, mixed, course, follow-up, and ambiguous turns.
Both nodes passed 306 focused Python tests and the documented live probes
before activation. This supports the routing implementation; it does not
retroactively turn the earlier conversation runs into tests of that extension,
or establish faculty-rated usefulness.

\subsection{Quizzes}

Both nodes passed full artifact and grading audits for all 65 modules. The
artifact audit bound each question to its current source excerpt and original
file hashes, then exercised all 435 questions, including deliberately wrong
answers. It checked that hints do not reveal keys or advance progress, and
that requests cannot carry another course's quiz state into the selected
offering. These checks establish reproducible practice behavior.

A separate 166-request live API audit included all 65 module starts on each
node, plus warm repeats, hints, and graded answers. Warm medians were 4.8\,ms
on Node~A and 5.8\,ms on Node~B, measured server-local. They describe stored
quiz operations and must not be compared as if they were generative tutoring
latencies or browser response times. Instructor review of answer correctness,
distractors, difficulty, and instructional coverage remains incomplete.

\section{Discussion}
\label{sec:discussion}

\subsection{Why retain the 8B classroom generator?}

Retaining \texttt{llama3.1:8b} is a deployment decision under incomplete quality
evidence, rather than a claim that eight billion parameters is an optimal
model size. The first bake-off established that several plausible upgrades
increased full-answer latency substantially under the same retrieval and
output budget. The second round qualified that result: Qwen~7B ran at almost
the baseline speed, and Mistral Nemo~12B stayed inside the prescribed 30\%
latency allowance. Consequently, ``larger models are too slow'' does not
adequately explain the final choice. The production policy requires a clear
grounded-tutoring benefit in addition to acceptable speed. Passing twelve
keyword-and-citation checks, or producing slightly more complete answers on
that set, does not establish that benefit. The documented decision is to keep
the existing generator while faculty compare the alternatives on the quality
of their actual explanations.

This choice preserves a known serving configuration while acknowledging its
material weaknesses. The fidelity diagnostic reproduced incorrect accounting
claims, failure to correct an earlier pricing error, and attribution of an
equation to an exercise that did not state it. Those failures preclude treating
continued deployment as evidence that the 8B model has passed the proposed
claim-support target. The defensible rationale is narrower:
no evaluated replacement has yet demonstrated the complete combination of
quality, continuity, integration, and operational readiness required for
promotion. A reviewed Qwen~7B or Nemo~12B improvement could change that
decision without changing the architecture.

\subsection{A model is evaluated through its serving contract}

The separate structured-response experiment illustrates why parameter count
alone is a poor predictor of either speed or usefulness. With thinking
disabled, the 35B-A3B mixture-of-experts candidate had a lower warm median than
the 8B baseline on the twelve fixed-source cases. It corrected important
misconceptions that persisted in the baseline, but also overstated the
protection provided by holding cash and refused an answerable reference to a
previous example. These are consequential tutoring regressions: a student
needs an explanation to preserve both the source's conditions and the setting
established in the conversation.

That comparison held the evidence and dialogue fixed but changed the model,
prompt, output schema, validation, and transport together. Its gains therefore
belong to the complete candidate configuration; they cannot be attributed
solely to model scale. Conversely, an empty response in the original
production-client bake-off does not show that the underlying model lacks
knowledge. The installed serving stack did not consistently apply the
structured format to thinking-capable chat requests when thinking was
disabled. A direct generation request resolved the observed JSON-formatting
problem, while factual and continuity failures remained. Runtime compatibility
thus belongs in the experimental specification alongside model tag,
quantization, context length, and sampling options. The experimental answer contract
also left production streaming and instructor-specific instructions
uncertified, providing another reason to withhold activation.

\subsection{Grounding requires useful answers and appropriate abstention}

The evidence suggests that stricter output validation trades some unsupported
answers for refusals, without eliminating semantic error. In the final
candidate, all twelve outputs were valid JSON and nine passed the quotation
and citation checks. An accepted answer nevertheless strengthened a
conditional financial benefit into a guarantee. Of the three withheld
answers, two lacked sufficient evidence and one was an answerable follow-up.
Exact quotation verifies a textual relationship with a source; it does not
verify that the explanation preserves the source's meaning.

The separate entailment experiment reinforces this distinction. Its proposed
contradiction threshold missed both original production errors, even though
the classifier detected simplified versions of some errors. A hypothetical
teaching example also received a neutral label, showing why rejecting all
non-entailed prose would suppress legitimate instruction.
A useful tutor must distinguish course-supported concepts, explicitly
invented example values, unavailable facts, and unjustified real-world
guarantees. Evaluation should therefore report unnecessary abstention and
loss of explanatory context alongside unsupported claims. The proposed
bounded revision experiment addresses these failure modes, but it remains
future work; neither a runtime factual guard nor a reliable semantic repair
mechanism has been demonstrated.

\subsection{Retrieval and conversation control are complementary to generation}

The multi-course resolver results identify a substantial source of improvement
outside the generator. Resolving a short follow-up into its intended retrieval
topic allows the existing model to receive relevant teaching material while
still answering the student's actual wording. Course and module boundaries
remain enforceable in this step. However, a correct topic does not ensure
that an adequate passage will be retrieved, and an adequate passage does not
ensure that the model will explain it faithfully. The five-course audit's
remaining retrieval misses, and the fixed-evidence fidelity failures,
demonstrate these distinct failure layers.

This decomposition helps allocate engineering effort. Clean-concept probes
recovered evidence for nineteen of twenty-four affected topics, indicating
that many misses arose from query formulation. Five topics still failed under
that favorable probe and required corpus, index, or threshold investigation.
A bounded expansion using only words in the student question recovered eleven
of the original queries without lowering the relevance floor. These findings
support targeted retrieval work; they do not imply that query processing can
repair missing material or generation errors. Preserving each turn's resolved
query, selected passages, and displayed answer makes that distinction
auditable and reduces the temptation to attribute every poor answer to model
size.

The subsequent conversation router, activated on both nodes on 20 September,
extends this separation to social, course, mixed, follow-up, and clarification
turns. Obvious cases use deterministic rules; ambiguous cases use a short
structured classification request to the same 8B model. Sharing the resident
model avoids introducing a second model and a possible model-loading penalty
while routing is evaluated. Its validated intent schema is a deployed control
mechanism, distinct from the experimental evidence-first answer contract
discussed above. The implementation also illustrates a deliberate complexity
trade-off: bounded browser-supplied history and a small Python router meet
the current requirements, while durable cross-device conversations would
require additional persistence. LangGraph remains an option for that future
requirement, rather than a dependency needed to obtain the reported
improvements.

\subsection{On-premises control carries operational obligations}

Local inference and separate offering collections give the institution
control over where course materials and student dialogue are processed and
which evidence enters an answer. Dialogue is handled on the campus hosts
rather than sent to an external generation API. Each offering is keyed by its
own CRN, so instructors can maintain different versions of the same catalog
course and decide which materials that version may retrieve. Sharing one
generator across offerings avoids maintaining six content-specific language
models, while separate collections, data trees, browser-held histories, and
request routing keep those CRN boundaries in place. Uploaded materials update
the retrieval collection for the selected offering; they are not used to
fine-tune a shared model on faculty or student content. Access is organized
around the LMS course context and course routing rather than an application-
level enrollment check in CourseChat itself. These mechanisms support
reproducible content updates and instructor-controlled corpora. Their
evaluation demonstrates scoped behavior on the recorded probes, rather than a
general proof of security, privacy compliance, or enrollment enforcement.

The two complete local stacks provide an operational alternative when one
node is under maintenance, but duplicate state introduces explicit
maintenance work. Updating the primary index is insufficient: the replica
must receive the corresponding snapshots, and quiz banks and evidence ledgers
must be copied separately. Code, model configuration, and course artifacts
also need consistent versions. This is the cost of retaining local ownership
and independently serviceable hosts. The reported work does not measure
capital amortization, electricity, administration time, or cost per student;
it therefore supports no claim that this deployment is cheaper than a cloud
service. Its demonstrated motivation is control under the institution's
chosen deployment constraints.

\subsection{Prebuilt practice moves uncertainty into a reviewable build stage}

Prebuilding quizzes changes both the computational and educational trade-off.
Serving a stored question and grading against its key avoids a generation
request for each practice turn, while making hints and explanations stable
across sessions. It also makes evidence and arithmetic review possible before
an item reaches a student. In exchange, the bank offers finite variation and
requires rebuilding or re-review when course material changes. Six or seven
questions per module establish availability across the module map, rather
than exhaustive coverage of each learning objective.

The distinction between artifact integrity and teaching quality remains
essential. Successful grading and exact source bindings demonstrate that
practice behaves consistently, but they cannot establish suitable difficulty,
plausible distractors, or a single pedagogically defensible answer. All banks
still record instructor review as incomplete. Similarly, a module with
limited source material can offer bounded, supported practice without
claiming comprehensive coverage. This explicit limitation is preferable to
letting live generation silently supply missing instructional content.

\subsection{Faculty review, student uptake, and educational value}

The next pilot stage should connect content maintenance to teaching practice.
Instructors could identify recurring misconceptions, contribute supplementary
cases or worked examples, and review the questions generated from those
materials. A useful approval workflow would present each question alongside
its answer, distractors, hint, explanation, and supporting excerpts, allowing
the instructor to edit, approve, or return it for revision. The existing
source bindings and item-review records provide an artifact foundation for
this workflow; a streamlined faculty interface and completed instructor
approval remain future work. Preserving approved edits and requesting renewed
review when their supporting material changes would make this feedback loop
part of maintaining a course, rather than a one-time inspection of generated
content.

Student uptake should be evaluated over a teaching period. Initial visits
would show reach, while return visits and continued use across later modules
would help distinguish sustained use from novelty. Reporting use separately
for explanation, conversational follow-up, and quiz practice would reveal
which activities students find valuable. Course coverage also matters: a
service used intensively in one offering may fit that instructor's materials
without transferring to other courses. These observations should be paired
with student feedback on usefulness, ease of use, trust in source-supported
answers, and reasons for abandoning an interaction. Acceptance cannot be
inferred from usage counts alone, particularly where an instructor explicitly
encourages participation.

Educational evaluation should examine what students can do after assistance.
A future study could combine interaction measures, such as attempts before
revealing answers and responses to hints, with delayed assessments and new
problems completed without the tutor. Appropriate comparisons include the
existing course resources and the general-purpose AI tools students already
use. Comparing equivalent learning tasks would help establish whether access
to instructor-selected evidence and reviewed practice adds value beyond
convenient text generation. Perceived usefulness, assisted task performance,
and independent learning should remain separate outcomes.

Faculty acceptance is equally consequential. The pilot should record the
time needed to prepare materials, review questions, resolve reported errors,
and update a course between offerings. A proposed feedback mechanism could
let students flag an unclear explanation or disputed answer for instructor
review, with confirmed issues feeding into corpus corrections and regression
cases. This would connect observed difficulties to concrete revisions while
making the teaching workload visible. Together, student and faculty evidence
would guide which interaction modes to extend and which content gaps deserve
attention first.

\subsection{Limits of the evidence and next evaluation priorities}

The reported latency measurements are predominantly serial and server-local.
Two API workers permit overlapping requests, but worker count and available
memory do not establish usable classroom throughput. Warm model timings also
exclude the substantial cold-load delay observed for the larger candidate.
The two-request overlap gate remains a requirement for promotion; the
presented results do not constitute a completed load study. Gateway
time-to-first-token requires separate measurement because proxy buffering can
delay visible output without changing generation time. SSE support is
implemented, whereas public gateway readiness and verification for every
course slug remained open in the documented deployment
snapshot.

The evaluation sets are engineering regressions rather than a representative
sample of student use. Repeated development on frozen fixtures can improve
those cases without transferring to unseen misconceptions or question forms.
Faculty ratings for usefulness and factual support remain pending, and no
learning-outcome or controlled classroom study is reported. The next evidence
needed is therefore paired faculty review on held-out dialogues, explicit
measurement of false refusals and source-condition errors, and concurrent
browser-to-gateway timing under representative load. These checks would make
a future model change defensible while preserving the current distinction
between an implemented system, measured engineering behavior, and demonstrated
educational benefit.

\section{Conclusion}
\label{sec:conclusion}

CourseChat began as a Dhillon School of Business pilot to provide useful local
learning support alongside the generative AI services already available to
students. Its contribution rests on the design of the whole teaching system:
separate CRN-scoped collections preserve instructional context and let
instructors control which materials each offering may retrieve, a traceable
ingestion pipeline connects responses to supplied materials, conversation
controls support follow-up questions, and prebuilt quizzes make practice and
feedback inspectable. These choices allow one shared generator to serve
different offerings while giving course content and instructional interaction
their own development and review processes. They also expose practical
trade-offs, including preparation effort, finite question-bank variety, and
the need to keep replicated course artifacts consistent.

The model trials inform one part of this design. We retain
\texttt{llama3.1:8b} because the evaluated alternatives have not yet demonstrated
an overall advantage across grounded explanation, conversational continuity,
latency, and serving compatibility. Credible 7B and 12B candidates remain
available for further assessment. The choice is therefore conditional on
evidence and can change without rebuilding the course architecture. The
observed retrieval and fidelity failures also show why improving a tutor
requires attention to evidence preparation and dialogue handling alongside
model capability.

The next development stage should extend the course ingestion pipeline with
relevant, reviewed material supplied by instructors: supplementary cases,
worked examples, lecture explanations, and contextual material beyond the
core assigned text. Better coverage can address missing evidence, but adding
documents alone does not ensure better answers; extraction quality, module
mapping, retrieval, and explanation fidelity must be evaluated together. A
shared, versioned corpus can support several interaction modes, including
course discussion, guided examples, concept comparison, and quiz practice,
without maintaining a separate knowledge base for each mode. These modes
would complement general-purpose assistants such as ChatGPT by centering
the instructor's selected context, expectations, and review of teaching
materials. The local service's purpose is to help students reason within
that context and apply it themselves.

Faculty participation should shape that development. A review workflow can
allow professors to inspect, revise, and approve quiz items and their supporting
evidence, while student reports of unclear or incorrect explanations can
inform subsequent content and evaluation updates. Existing review artifacts
provide a starting point; instructor approval of the current banks is still
pending. The next pilot should assess sustained student uptake, acceptance
of different interaction modes, and the workload this process creates for
faculty. Delayed and unassisted assessments, alongside comparisons with
existing resources and general-purpose assistants, would establish whether
the system supports independent learning. These findings should guide its
expansion toward an instructor-informed learning environment whose value
comes from local context, useful feedback, and students' own effort.

\section*{Acknowledgments}
We thank the University of Lethbridge Information Technology (IT) team for
its support of the pilot and its infrastructure, and colleagues at the
Dhillon School of Business for their contributions to the project.

\clearpage
\appendix
\section{Campus request path (deployment contract)}
\label{app:flow}

\begin{enumerate}[leftmargin=*]
  \item Student opens \texttt{https://tutor.example.edu/<slug>} through an LMS
        embed. The embed is the intended entry path; the slug is not advertised
        to students as a standalone bookmark, and enrollment enforcement is not
        claimed here.
  \item IIS rewrites the slug to \texttt{/?course=<id>} and proxies to Node~A or
        Node~B without requiring sticky session affinity.
  \item UI loads; student submits a turn with mode, optional chapter lock, and
        bounded browser-held history for the selected course.
  \item API resolves course config $\rightarrow$ collection. Quiz mode reads the
        bank; the conversation router selects scoped retrieval, social response,
        instant FAQ, or clarification according to the turn.
  \item The response returns through the gateway, optionally streamed.
        Course answers include citations confined to the selected course, with
        any module-scope fallback reported;
        social and clarification responses do not invent course citations.
\end{enumerate}

\end{document}